\documentclass[final,3p,times]{elsarticle}

\usepackage{amssymb}

\usepackage[ruled]{algorithm2e}
\usepackage{booktabs}
\usepackage{multirow}
\usepackage{amsmath}
\usepackage{url}
\usepackage{xcolor}

\begin{document}

\begin{frontmatter}



\title{GAPS: Geometry-Aware Problem Solver}


\author[label1]{Jiaxin Zhang\fnref{fn1}}
\ead{jiaxin.zhang@strath.ac.uk}

\author[label2]{Yinghui Jiang\fnref{fn2}}
\ead{yinghui@mindrank.ai}

\author[label1]{Yashar Moshfeghi\corref{cor1}}
\ead{yashar.moshfeghi@strath.ac.uk}

\cortext[cor1]{Corresponding author}
\fntext[fn1]{This is the first author footnote.}
\fntext[fn2]{XIAMEN University, China}

\affiliation[label1]{organization={University of Strathclyde},
            city={Glasgow},
            postcode={G1 2TF}, 
            country={UK}}

\affiliation[label2]{organization={Mind Rank Ltd.},
            city={HangZhou},
            country={China}}

\begin{abstract}
Geometry problem solving presents a formidable challenge within the NLP community. Existing approaches often rely on models designed for solving math word problems, neglecting the unique characteristics of geometry math problems. Additionally, the current research predominantly focuses on geometry calculation problems, while overlooking other essential aspects like proving. In this study, we address these limitations by proposing the Geometry-Aware Problem Solver (GAPS) model. GAPS is specifically designed to generate solution programs for geometry math problems of various types with the help of its unique problem-type classifier. To achieve this, GAPS treats the solution program as a composition of operators and operands, segregating their generation processes. Furthermore, we introduce the geometry elements enhancement method, which enhances the ability of GAPS to recognize geometry elements accurately. By leveraging these improvements, GAPS showcases remarkable performance in resolving geometry math problems. Our experiments conducted on the UniGeo dataset demonstrate the superiority of GAPS over the state-of-the-art model, Geoformer. Specifically, GAPS achieves an accuracy improvement of more than 5.3\% for calculation tasks and an impressive 41.1\% for proving tasks. Notably, GAPS achieves an impressive accuracy of 97.5\% on proving problems, representing a significant advancement in solving geometry proving tasks.

\end{abstract}








\end{frontmatter}


\section{Introduction}
\label{}
\begin{figure}[tbhp!] 
\centering
\includegraphics[width=0.7\textwidth]{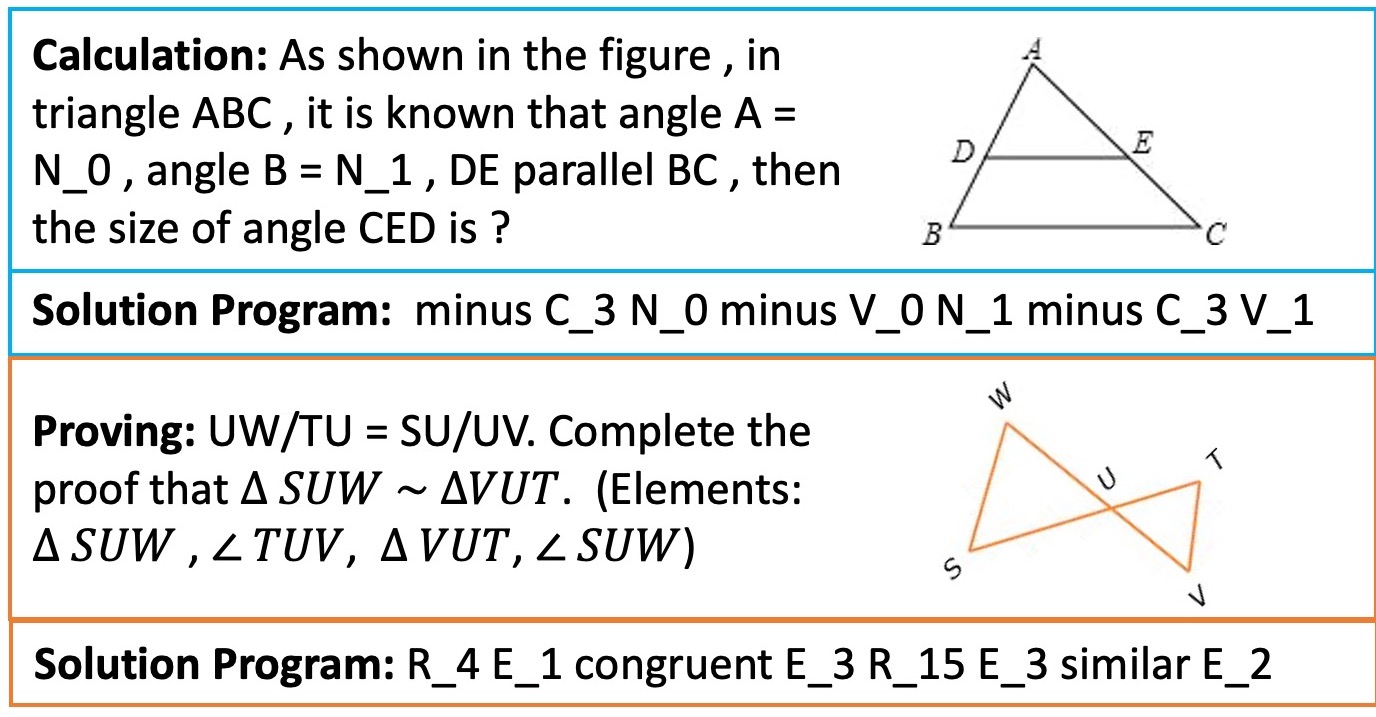} 
\caption{Two geometry math problems and related solution programs from the UniGeo dataset. Particularly, the problems in the blue rectangle box belongs to the calculation problem, whereas the problems in the orange rectangle box belongs to the proving problem. The operand ``C\_x" refers to the x-th constant, the operand ``N\_x" refers to the x-th numbers in the problem text, ``V\_x" refers to the results of the previous sub-program that at the x-th index of the total program, and the operand ``E\_x" refers to the x-th geometry element from the problem text. The operator "R\_x" refers to the x-th pre-defined theorem.}
\label{fig:geometry-problems}
\end{figure}

Deep learning techniques have shown remarkable success in automatically solving math word problems (MWPs) using methods such as CoT \cite{COT} and MultiHiertt \cite{MultiHiertt}. However, when it comes to automatically solving geometry math problems, the research is still at an early stage. Recent efforts have attempted to adapt models that succeeded in MWPs to handle geometry math problems, but only a few methods have yielded satisfactory results. One possible reason for this discrepancy is that, unlike MWPs, geometry math problems involve additional complexity due to the presence of geometric diagrams. To illustrate this point, consider the typical geometry math problem shown in Figure~\ref{fig:geometry-problems}. Effectively solving this problem requires a deep understanding of both the textual information and the accompanying diagrams \cite{GeoQA+}. Moreover, it is crucial to identify the geometry elements mentioned in the problem text to achieve satisfactory performance \cite{GEOS++}. 

Although researchers have been motivated to tackle geometry math problems from various angles, the existing approaches can be broadly categorized as utilizing pre-trained language models (PLMs) or recurrent neural networks to directly generate solution programs \cite{GeoQA, GeoQA+, UniGeo}. Consequently, these methods lack an inherent architectural design that explicitly captures the unique characteristics of geometry math problems.

Similar to research in the domain of mathematical reasoning, geometry math problems can be categorized into various types, with two dominant ones being calculation (CAL) problems and proving (PRV) problems (as depicted in Figure~\ref{fig:geometry-problems}). However, most recent works have primarily focused on CAL problems because other types of geometry math problems require different domain-specific languages. For example, solution programs for CAL problems involve arithmetic operators and numbers, while solution programs for PRV problems are constructed using geometric theorems and geometry elements. Nevertheless, it is essential to recognize that the mathematical knowledge required to solve CAL problems significantly overlaps with the knowledge needed for PRV problems \cite{UniGeo}. This observation underscores the considerable importance of developing a single solver that can generate solution programs for geometry math problems of all kinds of types. By creating such a versatile solver, we can leverage the shared mathematical knowledge and potentially devise more efficient and effective problem-solving techniques for a wide range of geometry math problems.

To resolve the difficulties in the automated geometry math problem solving, we propose the Geometry-Aware Problem Solver (GAPS), a model designed to automatically solve geometry math problems of different types simultaneously. GAPS leverages the VL-T5 \cite{VL-T5} model to obtain unified representations of text and diagrams, which are then used to predict the solution programs. To achieve a single solution program generator for various problem types, we assume that solution programs holds a unified pattern that consists of operators (arithmetic operators, geometric theorems) and operands (numbers, geometry elements). To handle different problem types, GAPS incorporates a problem-type classifier, which distinguishes between the problem types of various geometry math problems. By combining the unified solution program pattern with the problem-type classifier, GAPS can independently generate the operators and operands by selecting tokens from different domain-specific languages based on the specific problem types. Leveraging the unified solution program and the problem-type classifier, GAPS achieves the distinct capability of disentangling the process of generating the operators and operands within the solution program. To amplify this framework, we enhance GAPS by incorporating the hierarchical beam search, a decoding technique aligned with the approach of segregating operator and operand generation within GAPS. Furthermore, aiming to augment GAPS' proficiency in identifying geometry elements from the provided problem text, we introduce the geometry elements enhancement method which hinders the special geometry tokens being recognized as unknown tokens.

We assess the performance of our GAPS model using the recently published UniGeo dataset (UniGeo) \cite{UniGeo}, which comprises 4,998 problems involving calculations (CAL) and 9,543 problems requiring proofs (PRV). The evaluation of the GAPS model on the UniGeo dataset highlights its superiority over existing models. In comparison to the state-of-the-art Geoformer model \cite{UniGeo}, GAPS achieves significant accuracy improvements of 5.3\% for geometry calculation (CAL) problems and 41.1\% for proof (PRV) problems. Furthermore, we conduct experiments to explore the potential benefits of augmenting the GAPS model's training data. This augmentation is facilitated by GAPS' capability to solve diverse geometry problems using a single solution program generator, supported by the problem-type classifier. To elaborate, we utilize the recently introduced PGPS9K dataset \cite{pgps9k} that consists of 9,022 geometry math problems, out of which 2,891 problems are selected from another dataset called Geometry3K \cite{Inter-GPS}. We train the GAPS model using different combinations of datasets, namely the UniGeo dataset alone, the PGPS9K dataset alone, and a merged dataset containing both UniGeo and PGPS9K data. The results demonstrate that the combined dataset significantly enhances GAPS' performance. Additionally, we conduct an ablation study on the problem-type classifier, hierarchical beam search, and the geometry elements enhancement technology. The experimental outcomes underscore the substantial impact of these components in maximizing GAPS' ability to solve various types of geometry math problems. Overall, the comprehensive experimental results establish the efficacy and superiority of the GAPS model in addressing a wide array of geometry math problems.

The following are the key contributions of this paper:
\begin{enumerate}
    \item We present the GAPS model, which introduces a novel encoder capable of seamlessly processing both text and diagram inputs. Additionally, a geometry-specific program generator is incorporated to create solution programs for geometry math problems. This integration enables the model to effectively understand geometry diagrams.
    \item GAPS introduces a framework to handle a wide range of geometry math problems without requiring separate program generators for different problem types. This is achieved by inserting a problem-type classifier between the diagram and problem text encoder and the program decoder. This augmentation enhances GAPS' adaptability, allowing it to handle diverse types of geometry math problems.
    \item The geometry-specific program generator divides the generation of operators and operands in the solution program. This innovation presents a challenge for employing beam search decoding strategy. In this paper, we address this challenge by proposing hierarchical beam search as a pivotal enhancement for GAPS. This inventive beam search approach empowers GAPS to produce precise and varied solution programs for geometry math problems, leading to a significant performance boost.
    \item To comprehensively assess GAPS' capabilities, we conduct extensive experiments on various geometry math problem datasets, including UniGeo Calculation, UniGeo Proving, PGPS9K, and Geometry3K. The experimental outcomes underscore GAPS' exceptional performance compared to alternative methods designed for solving geometry math problems.
\end{enumerate}

\section{Related Work}

\subsection{Math Word Problems}
In recent years, the advancement of AI in solving mathematical word problems (MWPs) has been substantial. Researchers have introduced multiple high-quality datasets, including MathQA \cite{MathQA}, SVAMP \cite{SVAMP}, ASDiv \cite{ASDiv}, GSM8K \cite{GSM8K}, and TabMWP \cite{TabMWP}, each focusing on different aspects of MWPs. Moreover, the latest breakthroughs have been achieved by large pre-trained models using in-context learning methods, exemplified by GPT-3 and COT \cite{GPT-3, COT}, which have demonstrated remarkable proficiency in tackling MWPs. Additionally, models like ELASTIC \cite{ELASTIC} and MultiHiertt \cite{MultiHiertt} have made notable strides in enhancing mathematical reasoning capabilities. These collective efforts signify significant progress in the field of AI-based MWP solving and offer promising prospects for further research and applications.

\subsection{Geometry Math Problems Datasets}

In contrast to mathematical word problems (MWPs), the availability of high-quality datasets for geometry math problems has been limited. Previous datasets for geometry math problems either suffer from small scale or lack public accessibility \cite{GEOS++, GEO-OS, GeoShader}. However, in recent years, there has been progress, and several publicly available datasets, such as GeoQA  \cite{GeoQA}, GeoQA+ \cite{GeoQA, GeoQA+}, Geometry3K \cite{Inter-GPS}, and PGPS9K \cite{pgps9k}, have been released. Unfortunately, these datasets only cater to one type of geometry math problems, i.e. calculation problems.

To address the need for a comprehensive research interface covering both geometry calculation problems and proving problems, the UniGeo dataset \cite{UniGeo} has been introduced. This dataset is unique in its provision of both problem types and serves as a benchmark for researchers in the field. As of now, to the best of the authors' knowledge, UniGeo stands as the only available benchmark containing both geometry calculation problems and proving problems, enabling further advancements and evaluations in the domain of geometry problem-solving.

\subsection{Geometry Math Problems Solvers}

In the past, several works \cite{old-1, old-2, old-3, old-4} have been introduced to solve geometry math problems, but they were proposed long ago. In recent times, with the introduction of various public datasets, several neural-based methods have been proposed to tackle geometry problems. Examples include Inter-GPS \cite{Inter-GPS}, NGS \cite{GeoQA}, and DPE-NGS \cite{GeoQA+}. However, these methods have a significant limitation in effectively combining multiple modalities. For instance, DPE-NGS directly concatenates the diagram and text embeddings, leading to suboptimal fusion of the two modalities. Additionally, these methods do not address an important geometry problem type: PRV problems. This creates a motivation for further research in developing neural-based solvers tailored specifically for the geometry domain.

The Geoformer model \cite{UniGeo} endeavors to meet the challenge of uniformly encoding diagrams and texts by infusing diagram embeddings into the transformer as supplementary input. While it demonstrates commendable performance in both geometric calculations and proving scenarios, it neglects critical attributes specific to geometric mathematical issues, like geometric elements. Alternative approaches attempt to utilize Graph to model complex relationships among geometric primitives. For example, GeoDRL \cite{GeoDRL} translates both question diagram and text into a geometry logic graph, while PGDPNet \cite{PGDPNet} addresses the geometric diagram in the context of the scene graph generation problem. Nevertheless, preceding efforts typically treat symbolic characters (e.g., \(\triangle ABC\) as separate letters, thereby obstructing the semantic understanding of the geometric primitive \cite{symbolic_characters}. To overcome the limitations of prior methods, we propose GAPS in this work. GAPS is designed to effectively handle multiple modalities and takes into account the exclusive features of geometry problems, particularly geometry elements, thereby aiming to provide a more comprehensive and advanced solution for geometry problem-solving.

\section{Approach}
\label{sec:approach}

\begin{figure}[tbht!] 
\centering
\includegraphics[width=1.0\textwidth]{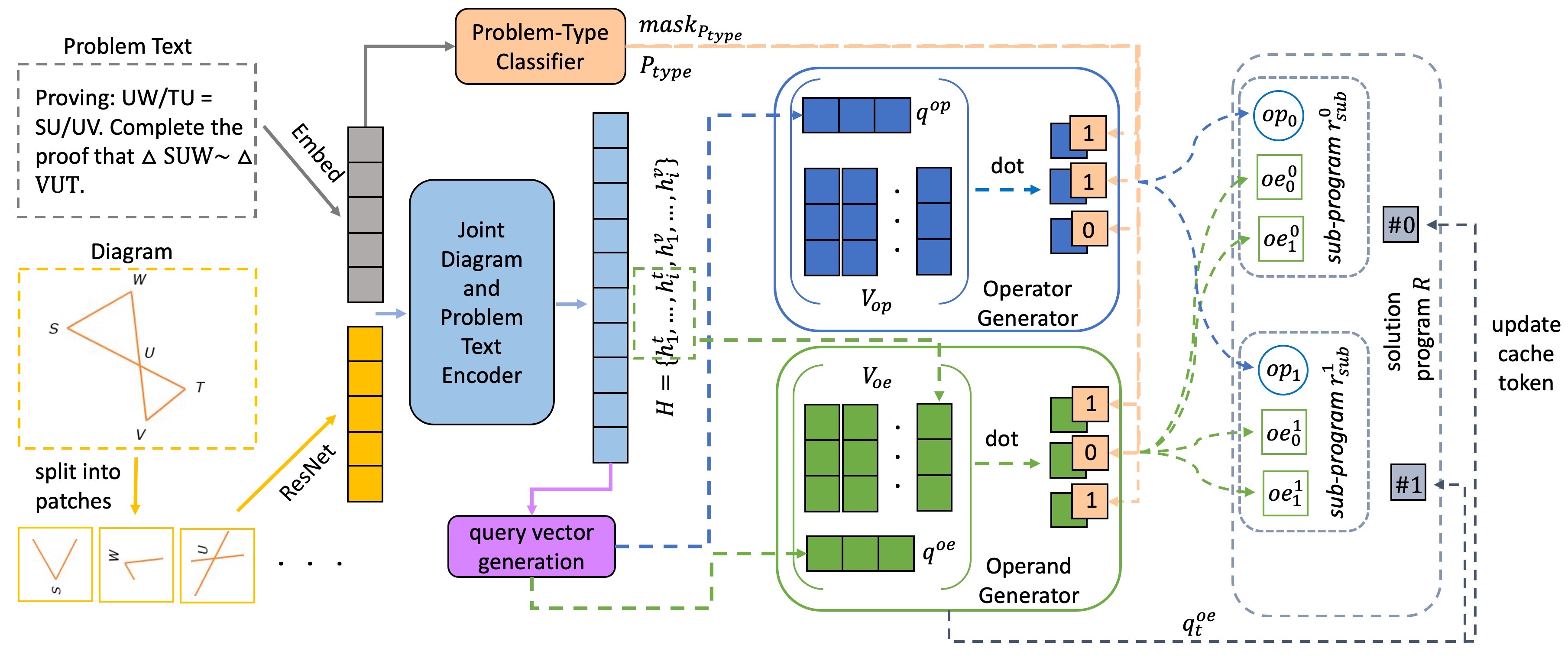} 
\caption{First, the diagram is divided into patches, and along with the problem text, both are transformed into vectors. These vectors are then concatenated and passed to the joint diagram and problem encoder, producing the joint representations denoted as $H$. Subsequently, the geometry-specific program generator starts to generate the solution program, each solution program is represented by several sub-program \(r^{i}_{sub}\). Each sub-program \(r^{i}_{sub}\) contains one operator \(op_{i}\) and corresponding operands \(\{oe^{i}\}\), where the query vector \(q^{op}\) is used for generating the operator \(op^{i}\), and the query vector \(q^{oe}\) is used for generating operands in the sub-program. After each sub-program is generated, the corresponding cache token is updated by replacing its vector by the \(q^{oe}_{t}\), which is the query vector used for generating the last operand in the sub-program. Specifically, $P_{type}$ is used to produce the mask to multiply with the probabilities of operators and operands, thus influencing GAPS to select $op$ and $oe$ from the domain-specific language (DSL) of the corresponding problem type. This mechanism enables GAPS to adapt and solve different kinds of geometry math problems effectively.}
\label{fig:model_architecture}
\end{figure}

We present a novel approach called the Geometry-Aware Problem Solver (GAPS) that is tailored specifically for solving geometry math problems. The architecture of our model is illustrated in Figure~\ref{fig:model_architecture}. GAPS employs a joint diagram and problem text encoder to create unified representations for both textual and diagram modalities. These unified text-diagram representations are then utilized by the geometry-specific program generator to produce the solution program, with a separation of the operators and operands. To enhance the accuracy of the generated solution programs, we introduce a hierarchical beam search for the geometry-specific program generator. Moreover, to handle various types of geometry math problems, we incorporate a problem-type classifier within GAPS, which enables the use of a single geometry-specific program generator to produce solution programs for different problem types. The generator employs masks according to the problem-type classifier to select the appropriate symbols from the domain-specific language defined for each type of geometry math problem.

\subsection{Task Definition}

The objective of our task is to generate the correct solution program \(\mathcal{R}\) for a given geometry math problem \(\mathcal{P}\) accompanied by a geometry diagram \(\mathcal{D}\). Each problem text \(\mathcal{P}\) belongs to a specific problem type \(P_{\text{type}}\). The solution program \(\mathcal{R}\) consists of both operators \(op\) and operands \(oe\). The operator \(op\) comprises various arithmetic operations (e.g., \(+, -\)) as well as advanced functions such as trigonometric functions and formulas for calculating areas. On the other hand, the operands \(oe\) can be pre-defined constant numbers or numbers (\(num\)) and geometry elements (\(ele\)) extracted from the problem text \(\mathcal{P}\). Please refer to Table~\ref{tab:terminology} for a detailed explanation of the notation symbols used in the context.

\begin{table}[tbhp!]
\centering
\caption{The notations used for the task definition.}
\begin{tabular}{@{}lp{8cm}}
\toprule
\textbf{Notation}  & \textbf{Description}                        \\ \midrule
\(\mathcal{P}\) & geometry math problem text \\
\(\mathcal{D}\) & geometry diagram \\
\(\mathcal{R}\) & solution program        \\ 
\(P_{\text{type}}\)       & problem type            \\ \midrule
\(r^{t}_{sub}\)      & the \(t\)-th sub-program of   \(\mathcal{R}\)          \\ 
\(op_{t}\)        & the operator in \(r^{t}_{sub}\)   \\
\(oe^{t}_{i}\) & the \(i\)-th operand in \(r^{t}_{sub}\)         \\ \midrule
\(\text{DSL}_{op}\)        & domain-specific language defined for operator           \\
\(\text{DSL}_{oe}\)        & domain-specific language defined for operand            \\ \midrule
\(\#t\)          &  cache token refers to  \(r^{t}_{sub}\)\\ 
\(num\)       & numbers from \(\mathcal{P}\)  \\
\(ele\)       & geometry elements from \(\mathcal{P}\) \\
\midrule
\end{tabular}
\label{tab:terminology}
\end{table}

\subsubsection{Normalize the Representation of the Solution Program}

\begin{figure}[tbht!] 
\centering
\includegraphics[width=0.6\textwidth]{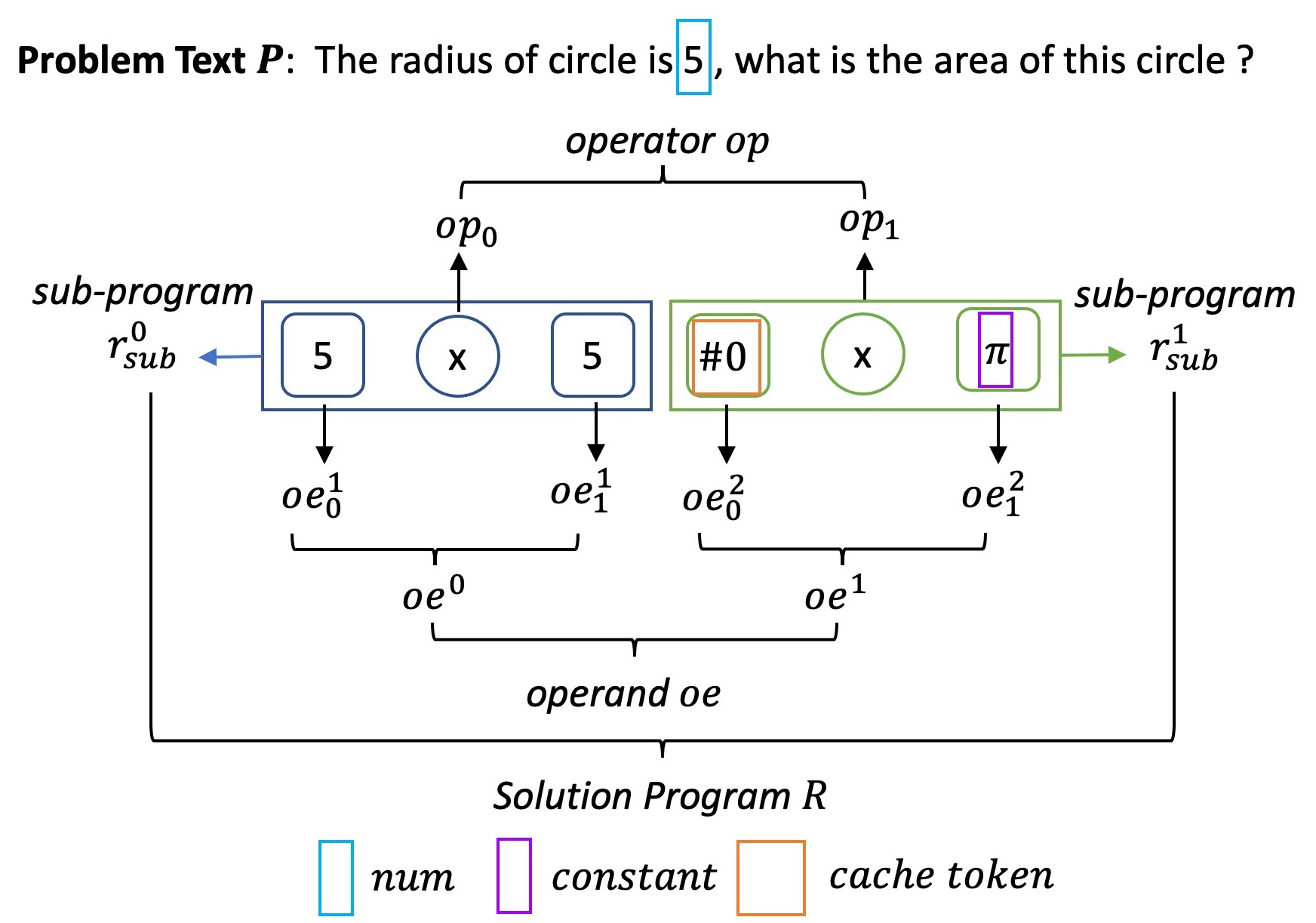} 
\caption{An example of normalized representation of the solution program.}
\label{fig:program_representation}
\end{figure}

We represent the solution program \(\mathcal{R}\) as a combination of operators and operands, adhering to the output format of GAPS's geometry-specific program generator. Moreover, the solution program \(\mathcal{R}\) is divided into a set of sub-programs \(r_{sub}\), each containing one operator and its corresponding operands. Formally, a solution program \(\mathcal{R}\) is defined as \(\mathcal{R}:= \{r^{t}_{\text{sub}}\}^{T}_{t=0}\), where \(T\) represents the total number of sub-programs in the solution program \(\mathcal{R}\), and \(r^{t}_{\text{sub}}\) refers to the \(t\)-th sub-program within the solution program. Furthermore, each sub-program \(r^{t}_{\text{sub}}\) is defined as \(r^{t}_{\text{sub}} := (op_{t}, \{oe^{t}_{i}\}^{I}_{i=0})\), where \(op_{t}\) denotes the \(t\)-th operator in the solution program \(\mathcal{R}\), and it is the sole operator included in the sub-program \(r^{t}_{\text{sub}}\). The set of \(\{oe^{t}_{i}\}^{I}_{i=0}\) is associated with \(op_{t}\), comprising a total count of \(I\).  For a concrete example, please refer to Figure~\ref{fig:program_representation}.

\subsubsection{Domain Specific Language}

As stated earlier, the geometry math problem text \(\mathcal{P}\) lack explicit domain-specific knowledge regarding operators and operands, making it necessary to define a domain-specific language (DSL) to aid the GAPS model in decoding. In particular, the DSL for operators (\(\text{DSL}_{op}\)) encompasses basic arithmetic operations, geometry theorems (e.g., Corresponding Angles Theorem), and geometry calculations (e.g., Circle\_R\_Area). Additionally, the DSL for operands (\(\text{DSL}_{oe}\)) includes constant numbers (e.g., the value of \(\pi\)) and cache token \(\#i\) that represents the result of the previous sub-program. For a comprehensive list of the DSL, please refer to the original dataset or the code link provided by us.

\subsection{Joint Diagram and Problem Text Encoder}

We opt for VL-T5 \cite{VL-T5} as our encoder due to its capability to uniformly embed both text and diagram modalities without requiring any modifications to the transformer architecture. Specifically, the geometry problem text \(\mathcal{P}\) is tokenized into tokens \(\{t_1, ..., t_i\}\) and mapped into learned text embeddings \(\{e^t_1, ..., e^t_i\}\), where \(i\) is the number of the tokens in \(\mathcal{P}\). Additionally, the diagram \(\mathcal{D}\) is divided into \(n\) patches, each of which is embedded into visual embeddings \(\{e^v_1, ..., e^v_n\}\). To create the joint representations, the text embeddings \(\{e^t_1, ..., e^t_i\}\) and visual embeddings \(\{e^v_1, ..., e^v_n\}\) are concatenated and fed into the VL-T5 transformer. The output of this process is the contextualized joint representations \(H=\{h^t_{1},...,h^{t}_{i},h^{v}_{1},...h^{v}_{n}\}\), where \(H \in \mathbb{R}^{(i+n) \times h}\), and \(h\) represents the dimension of the embeddings.

\subsubsection{Geometry Elements Enhancement}
The tokenizer employed by VL-T5 often converts geometry elements (e.g., \( \triangle \)) in the problem texts to unknown tokens, making it challenging for the program generator to differentiate between elements that share the same points (e.g., \( \angle \mathit{SUW} \) and \( \triangle \mathit{SUW} \)). To address this issue, we propose replacing geometry elements (e.g., \( \triangle \)) with geometry terminology tokens (e.g., "triangle"). Moreover, we introduce a straightforward yet useful technique called "geometry elements enhancement" to enhance the accuracy of selecting geometry elements \(ele\) from the problem text as operands. Specifically, we repeat and append these geometry terminology tokens to the problem text \(\mathcal{P}\). By doing so, the model becomes capable of distinguishing such elements accurately. In our ablation study, we demonstrate the significant effectiveness of this enhancement technology in improving the model's performance.

\subsection{Problem-Type Classifier}

In order to handle different types of geometry math problems with varying DSLs, it becomes necessary to combine the distinct DSLs into a unified one. However, this amalgamation unavoidably increases the complexity of solving geometry math problems, as it expands the search space of the decoding vocabulary. To address this challenge and alleviate the difficulty of generating the solution programs, we draw inspiration from the observation that neural models find it easier to perform discriminative tasks compared to generative tasks. As a solution, we introduce the problem-type classifier, which serves to distinguish between problem types for different geometry math problems. The problem-type classifier takes the contextualized representations \(H=\{h^t_{1},...,h^{t}_{i}\}\) of the problem text as input and subsequently generates probability distributions for all problem types:
\begin{equation}
\mathbb{P}(P_{\text{type}} | h^t_{1},...,h^{t}_{i}) = \text{softmax}(W_1 \cdot \text{sum}(h^t_{1},...,h^{t}_{i}, \text{dim}=1))
\label{eq:problem_classifier}
\end{equation}
where \(W_1 \in \mathbb{R}^{c \times h}\) is trainable parameters, an \(c\) is the number of problem types. Following the classification result \(P_{\text{type}}\), the problem-type classifier generates a mask that converts the probabilities of operators and operands not belonging to \(P_{\text{type}}\) into zeros. This masking mechanism ensures that irrelevant symbols are not generated, thus preventing the introduction of erroneous symbols in the solution program. By leveraging the problem-type classifier, GAPS can effectively address geometry math problems of various problem types with a single program generator. This approach significantly simplifies the training process as there is no need to separately train multiple program generators for different problem types, streamlining the model and promoting its versatility across different geometry math problems.

\subsection{Geometry-Specific Program Generator}

With the utilization of the joint diagram and problem text encoder, coupled with the problem-type classifier, the geometry-specific program generator takes the contextualized combined representations denoted as \(H\) as its input. It proceeds to construct the solution program \(\mathcal{R}\) by iteratively generating its sub-program \(r_{sub}\) over multiple steps, where each step is represented as \(i\). Building upon the insights from prior research \cite{ELASTIC}, which highlighted the advantages of segregating the generation process for operators and operands, our geometry-specific program generator adopts an alternating approach. Specifically, during the generation of each sub-program \(r^{i}_{sub}\) at the \(i\)-th step, the generator switches between generating the operators and the operands within the sub-program. This strategy empowers the GAPS model to proficiently devise accurate solution programs. It capitalizes on the contextual cues from both the problem text and diagram, while also drawing upon the problem-type classifier to ensure the appropriateness of the generated symbols with respect to the specific problem type.

\subsubsection{Generating Sub-Program Step-by-Step}

Given the inherent variability in decoding symbols \(s\) for numbers and geometry elements across different input geometry problems, it's important to note that each index within the decoding vocabulary doesn't consistently correspond to the same symbol \(s\). Moreover, the number of decoding options for both numbers and geometry elements within each input is contingent on the specific content of the original problem text. Consequently, the conventional approach of devising a fixed-length decoding vocabulary, commonly employed in language models, is unfeasible in this context.

In drawing inspiration from the principles of pointer networks \cite{pointer-network} and cross-attention architectures \cite{attention}, we introduce the concept of "decoding values vectors". This concept facilitates the program generator in selecting the appropriate symbol \(s\). Concretely, for a given input geometry problem, the decoding values vectors encompass the representations of all decoding symbols. These vectors are formulated as \(V = W_v (e_s)\), with \(e_s\) representing the embedding vectors for each symbol. Notably, \(e_s\) encompasses embeddings for both operators and operands defined within the domain-specific language (DSL), and these embeddings are amenable to training. Additionally, \(e_s\) incorporates vectors corresponding to numbers or geometry elements present in the problem text of the input geometry problem. These are formulated as \(h^{t}_{i}\), with \(i\) denoting the positional index of a number or geometry element within the geometry math problem text \(\mathcal{P}\).

In order to accurately choose the appropriate symbol \(s\) as the output for the current step, we formulate a query vector \(q_{s_{t-1}}\) that encapsulates the information derived from both the previously generated symbol \(s_{t-1}\) and the combined representation \(H\):
\begin{equation}
    q^{op,oe}_{t}, h_{t} = \text{GRU}_{op,oe}(\text{ReLU}(W_2[P_{aware} : v_{s_{t-1}}]), h_{t-1})
\end{equation}
where \(W_2 \in \mathbb{R}^{h \times 2h}\) is trainable parameters, and \text{ReLU} is the activation function, and \(h_{t-1}\) is the hidden state of the GRU from the previous step (by default, \(h_0\) is initialized as zero vector). Since we separate the generation of operators and operands, two GRUs with different parameters are adopted. The \(v_{s_{t-1}}\) refers to the value vector of the previous generated symbol \(s_{t-1}\). Notably, \(P_{aware}\) encodes the weighed sum information of \(H\), which is calculated by:
\begin{equation}
    P_{aware} = \sum_{i} a_i h^{t,v}_{i}
\end{equation}
where \(h^{t,v}_{i}\) is the representation vector of the \(i\)-th token in \(H\), and \(a_i\) is the weight, which is calculated through:
\begin{equation}
    a_i = \frac{\text{exp}(\text{score}(v_{s_{t-1}}, h^{t,v}_{i}))}{\sum_{L}\text{exp}(\text{score}(v_{s_{t-1}}, h^{t,v}_{j}))} 
\end{equation}
\begin{equation}
    \text{score}(v_{s_{t-1}}, h^{t,v}_{i}) = (v_{s_{t-1}}^{T}W_3) \cdot (W_4 h^{t,v}_{i})
\end{equation}
where \(W_3 \in \mathbb{R}^{h \times h}\) and  \(W_4 \in \mathbb{R}^{h \times h}\) are trainable parameters, and \(L\) is the total number of problem text tokens and diagram patches.

Subsequently, for the generation of the \(t\)-th symbol \(s_{t}\) within the sub-program \(r^i_{sub}\), we employ the query vector \(q^{op,oe}_{t}\) to perform a dot product with the decoding values vectors \(V\). This operation yields a probability distribution across all potential decoding candidates:
\begin{equation}
\label{eq:probability}
    \mathbb{P} = \text{softmax}(q^{op,oe}_{t} \cdot (W_5 \cdot V_{op,oe})) \times \text{mask}_{P_{type}}
\end{equation}
where \(W_{5} \in \mathbb{R}^{h \times h}\) is trainable parameters. The symbol with the highest probability among \(\mathbb{P}\) is selected as the predicted operator for the current step. When generating the initial symbol \(s_{0}\), since each sub-program comprises a single operator alongside multiple operands, the decoding value vector is limited to \(V_{op}\). This vector exclusively contains operators as defined within the DSL, facilitating the generation of \(op_{i}\).

Subsequent to generating the operator \(op_{i}\) for the sub-program \(r^{i}_{sub}\), GAPS replaces the decoding values vectors to \(V_{oe}\), which exclusively includes the constants specified within the DSL, as well as numbers (or geometry elements) present within the problem text. This adjustment allows GAPS to proceed with the completion of the associated operands \(\{oe^{i}\}\) within the sub-program \(r^{i}_{sub}\). The generation of operands concludes when the "end of sequence" token (denoted as ``\(eos\)"") is produced.

The sub-program \(r^{i}_{sub}\) is considered finished when both its operator \(op^{i}\) and operands \(\{oe^{i}\}\) have been generated. Following this, the program generator iterates through the same process outlined above to generate the following sub-programs \(r^{t > i}_{sub}\).

\subsubsection{Using one Program Generator for Different Problem Types}

In order to streamline the process and avoid the need for distinct program generators for varying problem types in geometry problems, we incorporate all symbols from the DSL into the decoding values vectors \(V\). To ensure that the chosen symbol adheres to the \(\text{DSL}_{P_{type}}\) specific to the given problem type \(P_{\text{type}}\), we utilize the mask \(\text{mask}_{P_{type}}\), generated by the problem-type classifier. This mask effectively eliminates probabilities associated with irrelevant symbols, ensuring that only valid symbols are taken into account during the generation process (as illustrated in Equation~\ref{eq:probability}). During training, the correct \(P_{\text{type}}\) information is provided by the ground truth. During the inference stage, the problem-type classifier (as seen in Equation~\ref{eq:problem_classifier}) is deployed to determine the \(P_{\text{type}}\) for the geometry math problem. This strategic approach empowers the GAPS model to employ a singular program generator capable of addressing diverse geometry math problems, thereby enhancing the efficiency of the overall architecture.

\subsubsection{Updating the Cache Token}

During the generation of sub-program \(r^{i}_{sub}\), its operands \(\{oe^{i}\}\) have the potential to be derived from the executable result obtained from the preceding sub-program \(r^{<i}_{sub}\). To enable the operand generator to effectively leverage results from earlier sub-programs, GAPS employs the cache token \(\#i\) within the \(\text{DSL}_{oe}\), drawing inspiration from previous work \cite{ELASTIC}. This cache token symbolizes the executable outcome of the \(i\)-th sub-program \(r^{i}_{sub}\). Diverging from other operands, the content encapsulated by \(\#i\) varies based on the specific sub-program it references. Consequently, GAPS needs to update its embeddings once the generation of sub-program \(r^{i}_{sub}\) is concluded.

To this end, we explore several approaches for updating the cache token \(\#i\):
\begin{itemize}
    \item Employing the \(q^{oe}_{t}\) as its embedding: \(q^{oe}_{t}\) signifies the query vector utilized in generating the last operand for sub-program \(r^{i}_{sub}\).
    \item Utilizing \(q^{op}_{0}\) as its embedding: \(q^{op}_{0}\) represents the query vector applied to generate the operator for sub-program \(r^{i}_{sub}\).
    \item Incorporating \(e_{op}\) as its embedding: \(e_{op}\) corresponds to the embedding of the operator generated for sub-program \(r^{i}_{sub}\).
\end{itemize}

By using the cache token \(\#i\), GAPS enhances its ability to make contextually informed decisions during operand generation, thereby improving overall performance.

\subsubsection{Hierarchical Beam Search for Operators and Operands}

\SetKwComment{Comment}{/* }{ */}
\begin{algorithm}[tbhp!] 
\caption{Hierarchical Beam Search}
\label{alg:beam_search}
\KwIn{$\textit{max}_{op}, \textit{max}_{oe}, bs, i,j=0, \textit{op}_{pred}=\text{[}\text{``sos"} \text{ for \_ in range } bs\text{]}, \textit{oe}_{pred}=\text{[}\text{ [ ] }\text{]}$}
\KwOut{$\mathcal{R}$}
\While{$i \neq \textit{max}_{op}$}{
        $\textit{ops}, \textit{ops}_\textit{scores} \leftarrow \text{Equation (6) }$  \Comment*[r]{\textit{ops}, $\textit{ops}_\textit{scores}$ are all predicted operands and probabilities, respectively.}
        
        $\textit{op}_\textit{candidates} \leftarrow \text{select\_max}(\textit{ops}, \textit{ops}_\textit{scores}, \textit{bs})$ \Comment*[r]{select the predicted operators of the beam size \textit{bs} with the maximum probabilities.}
        $\textit{beam}_\textit{sub\_oe} = \text{[ ]}$\;

        \For{$ \textit{op} $ \textbf{ in }  $\textit{op}_\textit{candidates}$}{
        
            $\textit{beam}_\textit{prev\_oe} = \text{[}\text{ [ ] } \text{ for \_ in range } \textit{max}_{oe}\text{]}$;

            \While{$j \neq \textit{max}_{oe}$}{
                $\textit{oes}, \textit{oes}_\textit{scores} \leftarrow \text{Equation (6) }$ \Comment*[r]{\textit{oes}, $\textit{oes}_\textit{scores}$ are all predicted operands and probabilities, respectively.}
                
                $\textit{oe}_\textit{candidates} \leftarrow \text{select\_max}(\textit{oes}, \textit{oes}_\textit{scores}, \textit{bs})$ \Comment*[r]{select the predicted operands of the beam size \textit{bs} with the maximum probabilities.}

                \If{$\textit{beam}_\textit{prev\_oe}[-1] \neq \text{none}$}{
                    $\textit{prev}_{oe\_ids} = \textit{oe}_\textit{candidates} // \textit{bs}$ \Comment*[r]{get the previous operands ids of the selected operands}

                    $\textit{beam}_\textit{prev\_oe} \leftarrow \textit{beam}_\textit{prev\_oe}[-1][\textit{prev}_{oe\_ids}]$ \Comment*[r]{save the previous operands of the selected operands}
                }\Else{
                    $\textit{beam}_\textit{prev\_oe} \leftarrow \textit{oe}_\textit{candidates}$
                }

                j +=1;
            }

            $\textit{beam}_\textit{sub\_oe} \leftarrow \text{backtrace}(\textit{beam}_\textit{prev\_oe}[-1])$ \Comment*[r]{backtrace to recover the selected operands sequences}

            $\textit{beam}_\textit{sub\_oe} \leftarrow \textit{oe}_\textit{candidates}$ \Comment*[r]{remember to append the last step prediction to the sequences}
        }

        $\textit{score}_\textit{op\_oe} \leftarrow \text{merge\_score}(\textit{op}_{candidates}, \textit{beam}_\textit{sub\_oe})$ \Comment*[r]{merge\_score: sum the probabilities of operators and their operands}

        $\textit{candidates}_\textit{op\_oe} \leftarrow \text{select\_max}(\textit{score}_\textit{op\_oe}, \textit{bs})$\ \Comment*[r]{selects beam size \textit{bs} of the combinations of operator and operands with highest probabilities.}

        $\textit{prev}_{op\_ids} \leftarrow \textit{candidates}_\textit{op\_oe} // \textit{bs}$ \Comment*[r]{get the previous operators ids of the selected sub-program}

        $\textit{op}_{pred} \leftarrow \textit{op}_{pred}\text{[}-1\text{]}\text{[}\textit{candidates}_\textit{op\_oe}\text{]} $ \Comment*[r]{save the previous operators of the selected sub-program}
        
        $\textit{oe}_{pred} \leftarrow \textit{beam}_\textit{sub\_oe}\text{[}\textit{prev}_{op\_ids}\text{]} $;

        i +=1;
}

$\{\textit{op}\}, \{\textit{op}_{index}\} \leftarrow \text{backtrace}(\textit{op}_{pred})$ \Comment*[r]{backtrace to recover the selected operator sequences}

$\{\textit{oe}\} \leftarrow \textit{oe}_{pred}\text{[}\{\textit{op}_{index}\}\text{]}$ \Comment*[r]{get the operands of the selected operator sequences}

$\{\textit{op}\} \leftarrow \textit{candidates}_\textit{op\_oe}.op$ \Comment*[r]{remember to append the last step prediction to the sequences}

$\{\textit{oe}\} \leftarrow \textit{candidates}_\textit{op\_oe}.oe$\;

\Return{$ \mathcal{R} = \{\textit{op}\} \vee \{\textit{oe}\} $}

\end{algorithm}

To ensure the production of high-quality solution programs and mitigate the possibility of sub-optimal solutions, GAPS enhances its geometry-specific program generator with the hierarchical beam search. This beam search approach aligns well with the design of segregating generations between operators and operands. Here is an outline of how the hierarchical beam search operates (refer to Algorithm~\ref{alg:beam_search}):

Let ``\(\textit{ops}\)" represent the set of all possible operators, and ``\(\textit{bs}\)" represent the beam size. Let \(\mathcal{R}\) represent the final solution program.
\begin{itemize}
    \item \textbf{Operator Generation:} At each step of generating the sub-program, the hierarchical beam search selects the number of \(\textit{bs}\) operators from \(\textit{ops}\) as candidates. Let \(\textit{op}_{candidates}\) be the set of operator candidates and output probabilities.
    \item \textbf{Operand Generation:} For each operator candidate \(op \in \textit{op}_{candidates}\), the operand generator takes \(op\) as input and produces the number of \(bs\) sequences of operands. Let \(\textit{oe}_{candidates}\) represent the set of operand sequences generated for operator candidate \(op\).
    \item \textbf{Score Computation:} Once completing the generation of operands for each operator candidate, the output probabilities between operators and their corresponding operands are summed up to obtain scores for each combination. Let \(\textit{score}_{op\_oe}\) denote the scores of all operator candidates with their operand sequences. 
    \item \textbf{Candidate Selection:} The \(\textit{score}_{op\_oe}\) are sorted in descending order. The algorithm selects the top \(\textit{bs}\) operator candidates along with their corresponding operand sequences.
    \item \textbf{Iteration:} Repeat the above steps until the generation of the solution program \(\mathcal{R}\) is complete. In each iteration, new operator candidates and operand sequences are selected based on the scores computed in the previous iteration.
\end{itemize}

By using the hierarchical beam search, GAPS can effectively explore different combinations of operators and operands, leading to the selection of high-quality solution programs. This approach is instrumental in improving the overall performance and accuracy of the model in solving geometry math problems.

\subsection{Training Objective}
During the training phase, the GAPS model's parameters are updated by minimizing the combined negative log-likelihoods. This sum encompasses the negative log-likelihoods associated with several components, including the ground truth geometry math problem type \(P_{\text{type}}\), the ground truth operator at each step in the ground truth solution program, and the ground truth operand for each sub-program in the ground truth solution program.

\begin{equation}
\begin{aligned}
L = - \{\mathrm{log}\mathbb{P}(P_{\text{type}}|H) + \frac{1}{L_{prog}}\sum^{L_{prog}}_{i=0}\mathrm{log}\mathbb{P}(op_{i}) + \frac{1}{L_{prog}}\sum^{L_{prog}}_{i=0}\frac{1}{L_{sub}}\sum^{L_{sub}}_{j=0}\mathrm{log}\mathbb{P}(oe^{i}_{j})\}
\end{aligned}
\end{equation}
where \(L_{prog}\) and \(L_{sub}\) refer to the length of the golden solution program and the length of the golden sub-program. \(op_{i}\) is the \(i\)-th golden operator in the solution program, and \(oe^{i}_{j}\) is the \(j\)-th golden operand in the \(i\)-th sub-program of the solution program.

\section{Experimental Setup}

\subsection{Datasets}
We conducted an extensive evaluation of our model using the UniGeo dataset \cite{UniGeo}, the latest dataset specifically designed for assessing mathematical reasoning ability in geometry math problems. The UniGeo dataset is an extension of the GeoQA dataset \cite{GeoQA}, encompassing 4,998 CAL problems from GeoQA and an additional 9,543 new PRV problems. For each problem in the UniGeo dataset, annotated step-by-step solution programs are provided. The UniGeo dataset further categorizes each problem into one of eight sub-tasks, along with the corresponding number of problems in each sub-task (shown in parenthesis): Angle (2,737), Length (1,869), Other (392), Parallel (443), Triangle (3,035), Quadrangle (1,704), Congruent (2,808), and Similarity (1,553).

To maintain consistency with the authors of the UniGeo dataset, we followed their approach to split the dataset into training, validation, and test subsets, with respective ratios of 0.7, 0.15, and 0.15. This dataset serves as a comprehensive benchmark for evaluating our model's performance in tackling diverse geometry problem types and assessing its mathematical reasoning capabilities.

\subsection{Baselines}
\label{sec:baseline}

\begin{itemize}
    \item \textbf{No Solution Programs} we selected three methods known for their effectiveness in multi-modal reasoning tasks: FiLM \cite{FiLM}, RN \cite{RN}, and MCAN \cite{MCAN}. It is important to note that these baselines differ from our proposed approach in that they do not generate solution programs to solve geometry math problems. Instead, they treat the problem as a classification task.
    \item \textbf{Encode Text and Diagram Separately} we also compare our model with Seq2Prog \cite{MathQA} and BERT2Prog \cite{BERT2Prog}. Seq2Prog employs a GRU-based as text encoder, while BERT2Prog utilizes BERT as the text encoder. For the diagram enoder, they adopt the ResNet \cite{resnet}. Both Seq2Prog and BERT2Prog follow a similar approach of directly concatenating text embeddings and diagram embeddings as input to an LSTM decoder.
    \item \textbf{Geometry-Specific Models} We also compare the two previous models specified for solving geometric problems: NGS \cite{GeoQA} and Geoformer \cite{UniGeo}. NGS adopts an LSTM \cite{LSTM} for encoding text and ResNet-101 \cite{resnet} for diagram encoding. The model combines information from both modalities through a co-attention reasoning mechanism. On the other hand, Geoformer is developed based on VL-T5 \cite{VL-T5} and is pre-trained on math expressions to enhance its performance.
\end{itemize}

\subsection{Evaluation Metrics}
To maintain consistency with previous works \cite{GeoQA, UniGeo}, we employ top-10 accuracy as the evaluation metric for both CAL problems and PRV problems. Top-10 accuracy measures the percentage of the top 10 predicted solution programs that contain the ground truth program. By using these evaluation metrics, we can gauge how well our model performs in comparison to the previous state-of-the-art models, for solving both CAL problems and PRV problems. This comparison enables us to assess the model's proficiency in generating accurate solution programs for various geometry math problems.

\subsection{Implementation Details}
The experiment code was written in Pytorch \cite{Pytorch} and Huggingface \cite{HuggingFace}. We opted for the pre-trained T5-base embedding \cite{T5} to represent the tokens in the problem text, and we utilized ResNet-101 \cite{resnet} as the encoder for diagrams. In addition, the same diagram processing method as previous work \cite{GeoQA} was used, which changes the background colour to white and resizes each diagram to \(224 \times 224\). The model was trained on a single NVIDIA A100 GPU (40G VRAM) for 100 epochs, with a learning rate of 2e-4 and a batch size of 40. The model's weights were optimized by Adam optimizer \cite{Adam}. We used teacher-forcing during training, where the probabilities using the ground-truth symbols for different epochs are: 0.0 when the epoch is less than 10, 0.1 when the epoch is less than 20, 0.5 when the epoch is less than 30, 0.8 when the epoch is less than 40, 0.9 when the epoch is less than 100.

\section{Experimental Results and Discussion}

\begin{table}[tbhp!]
\centering
\begin{small}
\caption{The top-10 accuracy (\%) comparison between baselines and our proposed GAPS model. $\diamond$, $\dagger$, and $\ddagger$ refer to "No Solution Programs",  "Encode Text and Diagram Separately", and "geometry-specific Models", respectively. We report performances separately when training models on only CAL problems, only PRV problems, and the entire UniGeo problems (see column header "Data"). Numbers in bold refer to the highest accuracy in the specific data subset. In addition, we report the accuracy differences between GAPS and the previous state-of-the-art model (Geoformer) in the parenthesis. Except for the results of Seq2Prog are taken from \cite{GeoQA}, other baselines' results are taken from \cite{UniGeo}.}
\begin{tabular}{lcccccccccc}
\hline
\multicolumn{1}{c}{} & \multicolumn{1}{l}{} & \multicolumn{3}{c}{\textbf{CAL (\%)}} & \multicolumn{6}{c}{\textbf{PRV (\%)}}   \\ \cmidrule(r){3-5} \cmidrule(r){6-11}
\textbf{Methods}     & \textbf{Data}        & All        & Angle      & Length      & All  & Par. & Tri. & Qua. & Con. & Sim. \\ \hline \hline 
FiLM$\diamond$                 & CAL                  & 31.7       & 34.0       & 29.7        & -    & -    & -    & -    & -    & -    \\
RN$\diamond$                   & CAL                  & 38.0       & 42.8       & 32.5        & -    & -    & -    & -    & -    & -    \\
MCAN$\diamond$                 & CAL                  & 39.7       & 45.0       & 34.6        & -    & -    & -    & -    & -    & -    \\
Seq2Prog$\dagger$              & CAL                  & 54.2       & 66.4       & 38.5        & -    & -    & -    & -    & -    & -    \\
BERT2Prog$\dagger$             & CAL                  & 54.7       & 65.8       & 42.1        & -    & -    & -    & -    & -    & -    \\
NGS$\ddagger$                   & CAL                  & 56.9       & 71.5       & 49.1        & -    & -    & -    & -    & -    & -    \\
Geoformer$\ddagger$            & CAL                  & 60.3       & 71.5       & 49.1        & -    & -    & -    & -    & -    & -    \\
GAPS (Ours) &
  CAL &
  \begin{tabular}[c]{@{}c@{}}65.4 \\ \textit{(+5.1)}\end{tabular} &
  \begin{tabular}[c]{@{}c@{}}77.7 \\ \textit{(+6.2)}\end{tabular} &
  \begin{tabular}[c]{@{}c@{}}50.4 \\ \textit{(+1.3)}\end{tabular} &
  - &
  - &
  - &
  - &
  - &
  - \\ \hline \hline
BERT2Prog$\dagger$            & PRV                  & -          & -          & -           & 48.0 & 15.5 & 48.1 & 28.5 & 49.5 & 77.6 \\
NGS$\ddagger$                  & PRV                  & -          & -          & -           & 53.2 & 13.2 & 56.6 & 29.8 & 57.1 & 79.4 \\
Geoformer$\ddagger$            & PRV                  & -          & -          & -           & 55.7 & 13.2 & 56.6 & 29.8 & 57.1 & 79.4 \\
GAPS (Ours) &
  PRV &
  - &
  - &
  - &
  \textbf{\begin{tabular}[c]{@{}c@{}}97.8 \\ \textit{(+42.1)}\end{tabular}} &
  \textbf{\begin{tabular}[c]{@{}c@{}}86.1 \\ \textit{(+72.9)}\end{tabular}} &
  \begin{tabular}[c]{@{}c@{}}100.0 \\ \textit{(+43.4)}\end{tabular} &
  \begin{tabular}[c]{@{}c@{}}99.6 \\ \textit{(+69.8)}\end{tabular} &
  \textbf{\begin{tabular}[c]{@{}c@{}}99.5 \\ \textit{(42.4)}\end{tabular}} &
  \textbf{\begin{tabular}[c]{@{}c@{}}92.0 \\ \textit{(+12.6)}\end{tabular}} \\ \hline \hline
BERT2Prog$\dagger$             & UniGeo                  & 52.0       & 63.1       & 39.2        & 48.1 & 15.4 & 48.0 & 31.7 & 49.5 & 75.1 \\
NGS$\ddagger$                  & UniGeo                  & 51.9       & 63.6       & 38.8        & 47.4 & 11.2 & 46.9 & 31.3 & 48.3 & 77.6 \\
Geoformer$\ddagger$            & UniGeo                  & 62.5       & 75.5       & 48.8        & 56.4 & 19.4 & 69.4 & 20.4 & 60.3 & 75.0 \\
GAPS (Ours) &
  \multicolumn{1}{l}{UniGeo} &
  \textbf{\begin{tabular}[c]{@{}c@{}}67.8\\  \textit{(+5.3)}\end{tabular}} &
  \textbf{\begin{tabular}[c]{@{}c@{}}78.2 \\ \textit{(+2.7)}\end{tabular}} &
  \textbf{\begin{tabular}[c]{@{}c@{}}54.6 \\ \textit{(+5.8)}\end{tabular}} &
  \begin{tabular}[c]{@{}c@{}}97.5 \\ \textit{(+41.1)}\end{tabular} &
  \begin{tabular}[c]{@{}c@{}}85.5 \\ \textit{(+66.1)}\end{tabular} &
  \textbf{\begin{tabular}[c]{@{}c@{}}100.0 \\ \textit{(+30.6)}\end{tabular}} &
  \textbf{\begin{tabular}[c]{@{}c@{}}99.6 \\ \textit{(+79.2)}\end{tabular}} &
  \begin{tabular}[c]{@{}c@{}}99.1 \\ \textit{(+38.8)}\end{tabular} &
  \begin{tabular}[c]{@{}c@{}}90.7 \\ \textit{(+15.7)}\end{tabular} \\ \hline \hline
\end{tabular}
\label{tab:overall-results}
\end{small}
\end{table}

\subsection{Results when trained on both UniGeo CAL and PRV}
We present the comprehensive results of our proposed GAPS model and the baseline models in Table~\ref{tab:overall-results}. Notably, GAPS outperforms all models trained on the entire UniGeo dataset, achieving the highest accuracy scores. Particularly, GAPS demonstrates a remarkable increase of 5.3\% in accuracy on CAL problems and a substantial improvement of 41.1\% on PRV problems compared to Geoformer, the previous state-of-the-art method. This highlights GAPS' superiority in effectively solving geometry math problems. Furthermore, it is worth noting that both GAPS and Geoformer utilize VL-T5 as encoders. However, GAPS' programs generator exhibits superior performance in producing solution programs for geometry math problems, indicating its suitability for this specific task. Additionally, GAPS outperforms NGS by 15.9\% on CAL problems and an impressive 50.1\% on PRV problems. This showcases the significance of designing complex feature fusion techniques to enhance reasoning ability in geometric problem-solving tasks. Finally, the unsatisfactory performance of BERT2Prog emphasizes the essential role of step-by-step solution programs, reinforcing their indispensability in tackling geometry math problems effectively. Overall, the results affirm GAPS' outstanding performance and effectiveness in generating accurate and high-quality solution programs for geometry math problems, outperforming other state-of-the-art models in this domain.

\subsection{Results when solely trained on UniGeo CAL}
When evaluating models trained solely on CAL problems, GAPS continues to exhibit significant superiority over the selected baselines. Compared to FiLM, RN, and MCAN, GAPS achieves an accuracy approximately twice as high as theirs. This underscores the importance of designing a solution program generation approach tailored for solving geometry math problems specifically. Furthermore, GAPS outperforms Geoformer on all sub-tasks of CAL problems, a trend consistent with their performance when trained on the entire UniGeo dataset. Moreover, models that uniformly encode text and diagram information, such as GAPS and Geoformer, consistently achieve significantly better results compared to models that naively concatenate embeddings (Seq2Prog, BERT2Prog) or use co-attention mechanisms (NGS). This suggests that performance improvements are observed when using complex feature fusion techniques to align text and diagram information effectively. The findings emphasize the superior capabilities of GAPS in solving geometry math problems and the advantages of employing sophisticated fusion methods for achieving accurate and comprehensive reasoning in geometry math tasks.

\subsection{Results when solely trained on UniGeo PRV}
When evaluating models trained solely on PRV problems, GAPS consistently outperforms the baselines with competitive advantages. Particularly, GAPS achieves an impressive 42.1\% higher accuracy than the previous state-of-the-art method, Geoformer. Compared to GAPS trained on the entire UniGeo dataset, GAPS trained on PRV problems performs slightly better, with a 1.0\% accuracy increment. Remarkably, GAPS achieves an overall performance of nearly 100\% accuracy in PRV problems, highlighting its exceptional proficiency in solving this problem type. The notable difference of 29.7\% between GAPS' performance on CAL problems and PRV problems indicates that GAPS excels in solving the latter type of problem. In the upcoming section, we provide an in-depth analysis to understand the reasons behind the disparity in GAPS' performance across these two problem types.

\begin{figure}[tbhp!] 
\centering
\includegraphics[width=0.9\textwidth]{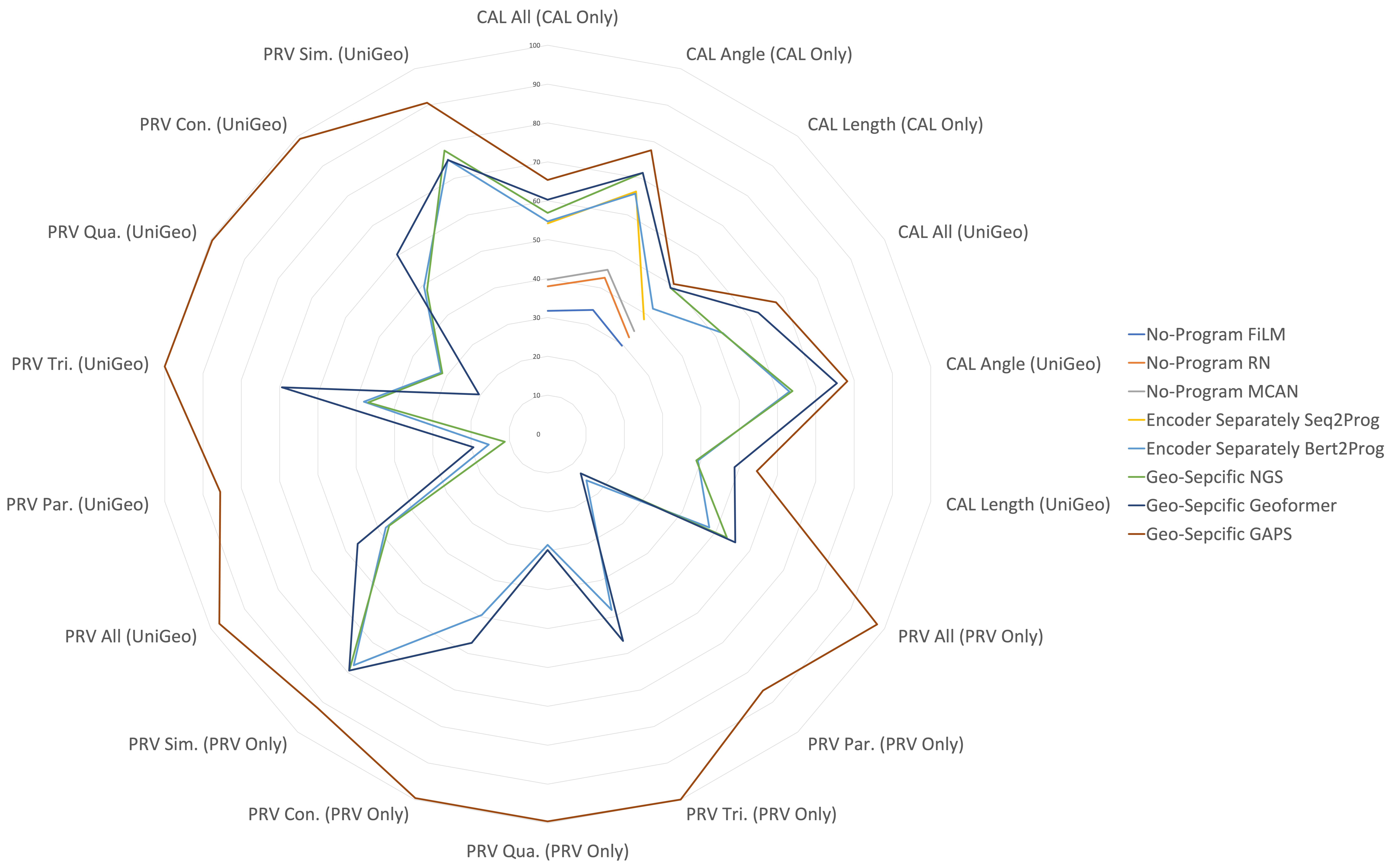} 
\caption{Overall results comparison in top-10 accuracy between GAPS and baselines on all sub-tasks by the radar chart.} 
\label{fig:radar_image}
\end{figure}

\subsection{Comparison of performances according to the sub-program types} 
Figure~\ref{fig:radar_image} presents a radar chart illustrating the performance of our proposed GAPS model and the baseline models in CAL problems, PRV problems, and their respective sub-tasks. The radar chart showcases the top-10 accuracy scores for each model in different problem types and sub-tasks. From the radar chart, it is evident that GAPS outperforms the baselines in both problem types and their corresponding sub-tasks in terms of top-10 accuracy. Notably, GAPS demonstrates significant advantages, particularly in the PRV problems.

\subsection{Performance Comparison with Geoformer Using Top-1 and Top-10 Accuracy Metrics}

\begin{table}[tbhp!]
\centering
\caption{Comparison between top-1 and top-10 accuracy (\%) on CAL problems and PRV problems achieved by Geoformer and GAPS, when training models on the entire UniGeo dataset.}
\begin{tabular}{lcccc}
\hline
                            & \multicolumn{2}{c}{CAL (\%)} & \multicolumn{2}{c}{PRV (\%)} \\ \cmidrule(r){2-3} \cmidrule(r){4-5}
\multicolumn{1}{c}{Methods} & top1          & top10        & top1          & top10         \\ \hline
Geoformer                   & 46.8         & 62.5         & 51.3          & 56.4         \\ \hline
GAPS (Ours)                         & 42.1         & 67.8      & 97.5        & 97.5        \\ \hline
\end{tabular}
\label{tab:top1vstop10}
\end{table}

Table~\ref{tab:top1vstop10} presents a comparison between our GAPS model and the previous best approach, Geoformer, using a stricter metric, top-1 accuracy, in addition to the top-10 accuracy. The table shows the top-1 and top-10 accuracy scores achieved by GAPS and Geoformer on CAL problems and PRV problems. Consistent with the results using top-10 accuracy, GAPS maintains its superiority over Geoformer on PRV problems when evaluated using top-1 accuracy. Specifically, GAPS achieves a remarkable 46.2\% improvement in top-1 accuracy compared to Geoformer, further highlighting its proficiency in solving PRV problems. However, when evaluating CAL problems using top-1 accuracy, our proposed GAPS model exhibits a decrease in performance of 4.7\% compared to Geoformer. This contrasts with the results obtained using top-10 accuracy, where GAPS outperforms Geoformer. This discrepancy suggests that GAPS may not always achieve the top predicted solution program in CAL problems, although it still excels in identifying the correct solution program among the top 10 predictions. The variation in performance on CAL problems under different evaluation metrics warrants further investigation in the subsequent analysis to gain a deeper understanding of GAPS' performance characteristics across different problem types.

\subsection{Analyzing Prediction Consistency: Top-1 and Top-10 Accuracy Comparison}

\begin{figure}[tbhp!] 
\centering
\includegraphics[width=1.0\textwidth]{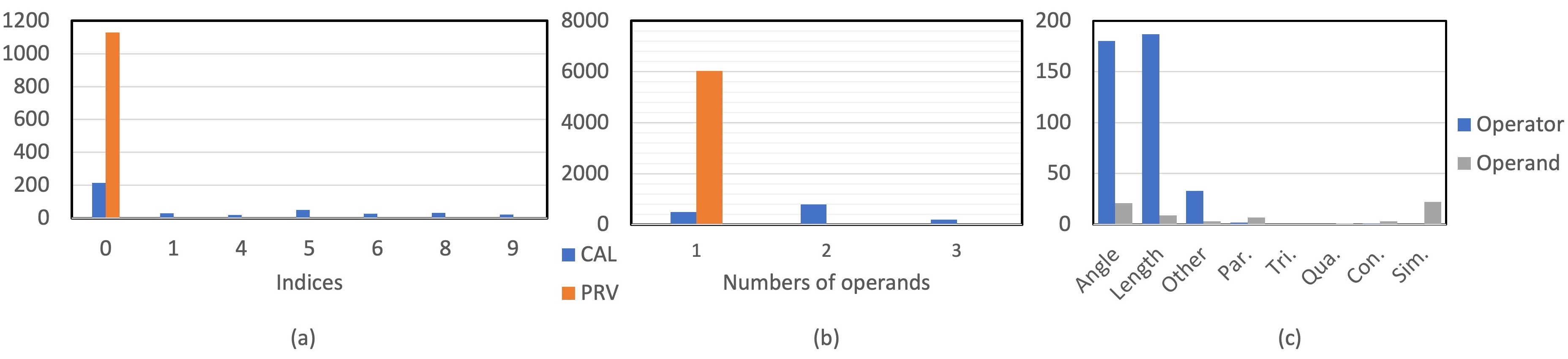} 
\caption{(a) Numbers of occurring indices of correct predictions in top-10 predictions candidates. We discard 2th, 3th, and 7th indices because numbers are zero. (b) Distributions of number of operands following each operator in ground truth solution programs from CAL problems and PRV problems. (c) The number of incorrect predictions due to wrong operators or wrong operands generated by GAPS in all sub-tasks. The "Angle", "Length", and "Other" belong to CAL problems, and the others belong to PRV problems.} 
\label{fig:top1-index}
\end{figure}

Table~\ref{tab:top1vstop10} reveals a significant discrepancy in the performance of GAPS on PRV problems compared to CAL problems, in terms of top-1 and top-10 accuracy. Interestingly, the top-1 and top-10 accuracy achieved by GAPS on PRV problems are identical, indicating that the model consistently makes correct predictions within the top 10 candidates for these types of problems. In contrast, on CAL problems, there is a considerable difference of 25.7\% between the top-1 and top-10 accuracy scores. To investigate the reasons behind this inconsistency, we conduct a detailed analysis by counting the occurrence of correct predictions among the top 10 candidates. Figure~\ref{fig:top1-index}(a) visually presents the distribution of correct prediction indices for both PRV problems and CAL problems. From Figure~\ref{fig:top1-index}(a), we make a significant observation: for PRV problems, all indices of correct predictions are clustered around the 0th index, indicating that the correct solution program is consistently ranked first among the top 10 predictions by GAPS. This finding provides an explanation for the identical top-1 and top-10 accuracy scores obtained by GAPS on PRV problems. In contrast, for CAL problems, the indices of correct predictions are dispersed across several positions within the top 10 predictions. This scattering of correct predictions leads to the difference between the top-1 and top-10 accuracy scores for CAL problems.

\subsection{Analyzing Performance Discrepancy: Complexity of CAL and PRV Problems}


Table~\ref{tab:overall-results} presents the top-10 accuracy comparison between the baselines and our proposed GAPS model. An interesting observation is the substantial performance difference in GAPS between CAL problems and PRV problems. This discrepancy could be attributed to the inherent complexity of CAL problems, which typically involve more intricate reasoning processes. To investigate this further, we refer to previous research \cite{ELASTIC} that highlights the significance of the number of operands following each operator in determining the complexity of solution programs. As such, we plot the distribution of the number of operands following each operator in the ground truth programs for both CAL and PRV problems Figure~\ref{fig:top1-index}(b). From Figure~\ref{fig:top1-index}(b), it is evident that all operators in PRV problems have only one following operand. Conversely, CAL problems exhibit greater variance in the number of operands, ranging from 1 to 3, per operator. This observation confirms our hypothesis that the complexity of solution programs is a key factor influencing GAPS' differing performances on the two problem types. PRV problems, with simpler solution program structures, are more effectively handled by GAPS, resulting in higher accuracy. However, the greater complexity of CAL problems poses a greater challenge, leading to a comparatively lower accuracy for this problem type. 

Additionally, we posit that generating correct operators poses a more challenging task than generating correct operands. The complexity of operator generation arises from the necessity to comprehend the semantic meaning of the text and diagrams while also manipulating relevant theorem knowledge. On the other hand, generating operands primarily requires matching entities and numbers in the text and employing finite domain knowledge, such as constants. In line with this hypothesis, we analyze and present the number of wrong predictions caused by mistaken operators or operands in Figure~\ref{fig:top1-index}(c). The findings reveal that in CAL problems, the majority of incorrect predictions are attributed to wrong operators. Conversely, in PRV problems, the incorrect predictions are more frequently due to wrong operands. This disparity in incorrect predictions further validates our hypothesis, demonstrating that generating solution programs for CAL problems is indeed more intricate than generating those for PRV problems. The complexity associated with CAL problems poses challenges for the model, leading GAPS to be more prone to fail in generating correct solution programs for this problem type.

Indeed, our analysis of the difficulty level between CAL problems and PRV problems clearly indicates that CAL problems pose a greater level of diversity and challenge in their solution programs. This inherent complexity in CAL problems explains the significantly better performance achieved by GAPS on the PRV problems than on the CAL problems.

\subsection{Dataset Augmentation with PGPS9K: A Comparative Analysis}

Due to the presence of the unified solution program pattern and the problem-type classifier, GAPS can effectively address a wide range of geometry math problems using just one solution program generator, all while maintaining the model's performance. Furthermore, our research seeks to determine whether enhancing the training data of GAPS with additional geometry math problems featuring diverse domain-specific languages (DSLs) could lead to performance improvements. Consequently, we conduct experiments to investigate the potential advantages of supplementing the GPAS model's training data with datasets that incorporate various DSLs alongside the UniGeo dataset. Specifically, we extend the data by incorporating the PGPS9K dataset \cite{pgps9k}, which represents the latest collection of geometry math problems sourced independently from the UniGeo dataset. The PGPS9K dataset consists of a total of 9,022 geometry math problems, out of which 2,891 problems are selected from another dataset called Geometry3K \cite{Inter-GPS}. Notably, the PGPS9K dataset encompasses 30 types of operators in its DSL, whereas the UniGeo dataset comprises 18 types of operators for CAL problems and 44 types of operators for PRV problems. To evaluate the GPAS model's performance, we use two distinct subsets of the PGPS9K test data: the first subset (Geometry3K test) includes 589 problems from the original Geometry3K dataset, and the second subset (PGPS9K test) includes 1000 problems sourced from the PGPS9K dataset.

\begin{table}[tbhp!]
\centering
\caption{The performance of top-10 accuracy (\%) achieved by GAPS model on various datasets, inlcuding UniGeo CAL, UniGeo PRV, PGPS9K test, and PGPS9K test datasets. The column displaying the highest accuracy is marked in bold. For datasets that were not used during the training stage, the accuracy is not reported, and the corresponding cells are marked as "n/a."}
\begin{tabular}{@{}lllll@{}}
\toprule
Training Dataset         & CAL (\%)          & PRV (\%)          & PGPS9K test (\%)       & Geometry3K test (\%)    \\ \midrule
UniGeo          & 67.8          & \textbf{97.5} & n/a           & n/a           \\ \midrule
PGPS9K          & n/a           & n/a           & 50.5          & 57.5          \\ \midrule
UniGeo + PGPS9K & \textbf{68.5} (\textit{+1.7}) & 97.2 (\textit{-0.3})         & \textbf{61.2} (\textit{+10.7}) & \textbf{68.0}  (\textit{+10.5}) \\ \bottomrule
\end{tabular}
\label{tab:add-more-data}
\end{table}

In Table~\ref{tab:add-more-data}, we present the varied performances of the GAPS model based on different training approaches. Specifically, we compare the model's performance when trained on three different datasets: UniGeo dataset alone, PGPS9K dataset alone, and the combined UniGeo and PGPS9K datasets. The results reveal a notable improvement in the GAPS model's performance on UniGeo CAL, PGPS9K, and Geometry3K datasets when trained on the combined dataset, in comparison to training with a single dataset. The accuracy shows an increase of 1.7\%, 10.7\%, and 10.5\% on UniGeo CAL, PGPS9K, and Geometry3K, respectively. Despite the substantial differences in the DSL between UniGeo and PGPS9K, the addition of the PGPS9K dataset during training still enhances the GAPS model's capability to solve geometry math problems. However, it is noteworthy that the GAPS model experiences a negligible decrease in performance on the UniGeo PRV test data when trained on both UniGeo and PGPS9K datasets. This can be attributed to the distinct problem types present in the PGPS9K dataset compared to the UniGeo PRV dataset. The former dataset primarily consists of calculation-type problems, while the latter focuses on proving-type problems.

In summary, the findings from Table~\ref{tab:add-more-data} demonstrate that augmenting the GAPS model's training data with datasets featuring different Domain-Specific Languages (DSLs) can actually enhance its performance, rather than causing any detriment. Interestingly, the design of GAPS, which separates the generation of operators and operands, plays a crucial role in mitigating the complexities introduced by the additional dataset. This design choice allows GAPS to effectively capitalize on the benefits offered by the new dataset without getting overwhelmed by its inherent complexity. As a result, GAPS demonstrates improved performance when trained on the combined dataset with a diverse DSL, underlining the advantages of incorporating data from different domains.

\subsection{Analysis of the Utility of the Problem-Type Classifier}

This section explores the significance of the problem-type classifier in GAPS, which enables the model to generate solution programs for geometry math problems belonging to different types simultaneously.

\begin{table}[tbph!]
\centering
\caption{The comparison between the top-10 accuracy (\%) achieved by GAPS model under three configurations: 1. without the problem-type classifier (denoted as "w/o"), 2. equipped with the problem-type classifier (denoted as "w"), 3. considering UniGeo CAL, UniGeo PRV, and PGPS9K as three different problem-types (denoted as "three types"). The highest accuracy scores in each row are highlighted in bold type, indicating the best performance achieved by the respective configuration for each problem type.}
\label{tab:problem-classifier}
\begin{tabular}{@{}llccc@{}}
\toprule
Train Data                       & Test Data  & w/o (\%) & w (\%)            & three types (\%) \\ \midrule
\multirow{2}{*}{UniGeo}          & CAL        & 46.8 & \textbf{67.8} & n/a           \\ \cmidrule(l){2-5} 
                                 & PRV        & 96.9 & \textbf{97.5} & n/a           \\ \midrule
\multirow{4}{*}{UniGeo + PGPS9K} & CAL        & 40.4 & \textbf{68.5} & 49.4          \\ \cmidrule(l){2-5} 
                                 & PRV        & 97.1 & 97.2          & \textbf{97.9} \\ \cmidrule(l){2-5} 
                                 & PGPS9K test     & 53.0 & \textbf{61.2} & 58.8          \\ \cmidrule(l){2-5} 
                                 & Geometry3K test & 61.0 & \textbf{68.0} & 63.5          \\ \bottomrule
\end{tabular}
\end{table}

\subsubsection{The Impact of Problem-Type Classifier on GAPS Performance}

The comparison between the performance of GAPS with and without using the problem-type classifier is illustrated in Table~\ref{tab:problem-classifier}. When GAPS is trained solely on the UniGeo dataset, the presence of the problem-type classifier leads to significantly improved performances. Specifically, GAPS equipped with the problem-type classifier achieves a substantial increase of 21\% in accuracy on the UniGeo CAL dataset and a 0.6\% increase on the UniGeo PRV dataset, compared to GAPS without it. Furthermore, when GAPS is trained on the combined UniGeo and PGPS9K datasets, the importance of the problem-type classifier becomes even more evident. GAPS without the problem-type classifier exhibits notably lower accuracy across all test data in comparison to GAPS with the classifier. These results presented in Table~\ref{tab:problem-classifier} emphasize the effectiveness of the problem-type classifier. By acting as a discriminative task, the problem-type classifier efficiently handles the complexity within the generation of solution programs, leading to enhanced performance for the GAPS model on various geometry math problem types.

\subsubsection{Evaluating the Influence of Fine-Grained Problem-Type Classification on GAPS Performance}

We initiated an inquiry into the potential benefits of categorizing geometry math problems into finer problem types. To explore this, we treated UniGeo CAL, UniGeo PRV, and PGPS9K as three distinct problem types. As depicted in Table~\ref{tab:problem-classifier}, when dividing the datasets into three problem types, the performance of GAPS exhibited a decline on UniGeo CAL, PGPS9K, and Geometry3K test data, with accuracy drops of 19.1\%, 2.3\%, and 4.5\%, respectively. Upon closer examination, we realized that problems within the UniGeo CAL, PGPS9K, and Geometry3K datasets predominantly focus on calculation problems, warranting their categorization into a single problem type. Despite the performance decrease in the finer-grained problem type classification, the accuracy of GAPS on these three test datasets remains substantially higher than GAPS without any problem-type classifier at all. From these observations, we can confidently conclude that GAPS indeed benefits significantly from the problem-type classifier, and the extent of its benefits is maximized when the correct problem types are appropriately identified.

\subsubsection{Accelerated Convergence with Problem-Type Classifier in GAPS Training}

\begin{figure}[tbph!] 
\centering
\includegraphics[width=1.0\textwidth]{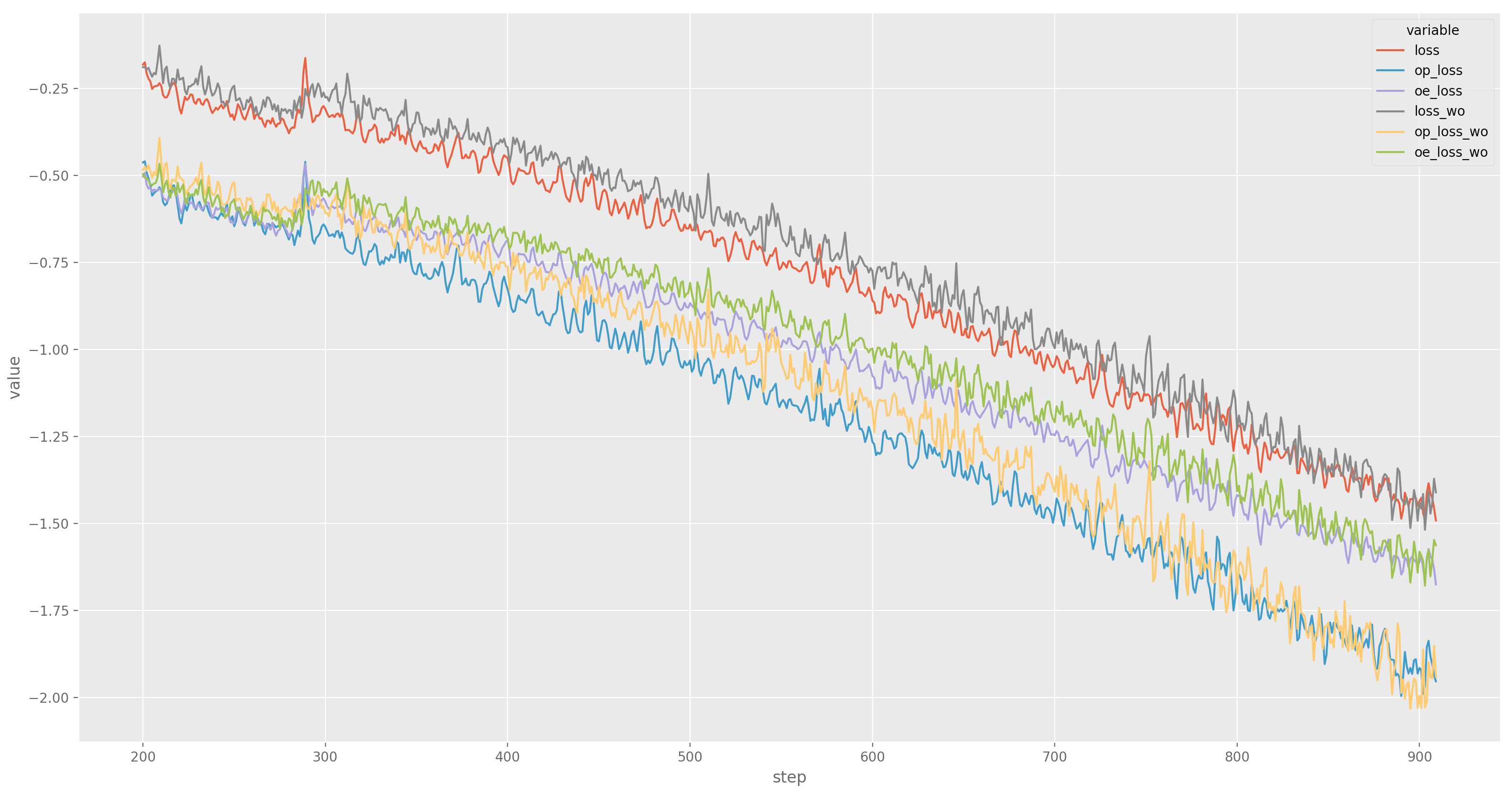} 
\caption{Comparison of training loss convergence in GAPS, with and without the utilization of the problem-type classifier. The ``loss" and ``loss\_wo" are the sum of operator and operands training losses values with and without problem-type classifier. The ``op\_loss" and ``op\_loss\_wo" are the operator training loss values with and without problem-type classifier. The ``oe\_loss" and ``oe\_loss\_wo" are the operand training loss values with and without problem-type classifier. The loss values are scaled by the logarithmic function for better visualization.} 
\label{fig:pc-loss}
\end{figure}

We have generated Figure~\ref{fig:pc-loss} to visualize the training losses of GAPS with and without the problem-type classifier. The plot illustrates that GAPS equipped with the problem-type classifier converges faster than GAPS without it. The reason behind this accelerated convergence lies in the problem-type classifier's role in simplifying the task of generating solution programs. By initially classifying the input geometry math problem, the problem-type classifier provides valuable insights, streamlining the subsequent process of generating solution programs for the GAPS model.

\subsection{Analysis for the Different Cache Token Updating Methods}

Table~\ref{tab:cache_token} presents a performance comparison of various strategies for updating cache tokens. As indicated in the table, employing the query vector used to generate the final operand in the sub-program to update the cache token achieves the highest accuracy across all datasets. Our hypothesis is that this outcome can be attributed to the comprehensive nature of the query vector \(q^{oe}_{t}\), which encapsulates information from all the symbols generated within the sub-program. Consequently, it can be interpreted as an integrated representation of the sub-program. In contrast, the other two vectors used to update the cache token, \(q^{op}_{0}\) and \(e_{op}\), do not take into account the generated operands, resulting in the incorporation of only partial information from the sub-program.

\begin{table}[tbhp!]
\centering
\caption{Performance comparison between different cache token updateing methods. \(q^{op}_{0}\) refers to the query vector used for generating the operator for the sub-program, \(e_{op}\) refers to the embedding of the generated operator for the sub-program, and \(q^{oe}_{t}\) refers to the query vector used for generating the last operand for the sub-program. For an equitable evaluation, GAPS models trained exclusively on UniGeo CAL, UniGeo PRV, and PGPS9K datasets are used to assess the performance on their respective test datasets.}
\label{tab:cache_token}
\begin{tabular}{@{}ccccc@{}}
\toprule
                      & CAL (\%)           & PRV  (\%)         & PGPS9K   (\%)     & Geometry3K (\%)   \\ \midrule
\(q^{op}_{0}\)    & 59.0          & 91.0          & 36.5          & 40.2          \\ \midrule
\(e_{op}\) & 58.8          & 90.2          & 37.6          & 42.2          \\ \midrule
\(q^{oe}_{t}\)    & \textbf{65.4} & \textbf{97.8} & \textbf{50.5} & \textbf{57.5} \\ \bottomrule
\end{tabular}
\end{table}

\subsection{Enhancing Solution Program Generation with Hierarchical Beam Search}

To enable GAPS to utilize beam search for generating solution programs, we propose the hierarchical beam search, which is designed to be compatible with GAPS' architecture that separates the generation of operators and operands. In order to showcase the necessity and effectiveness of the hierarchical beam search, we conduct a comparison between GAPS employing the hierarchical beam search and GAPS using the traditional greedy decode approach. The results are presented in Table~\ref{tab:hierarchical-beam-search}, which clearly demonstrates that GAPS with hierarchical beam search outperforms the GAPS with greedy decode by a significant margin.

\begin{table}[tbhp!]
\centering
\caption{Comparing Accuracy (\%) of GAPS with Hierarchical Beam Search and Greedy Decode for Solution Program Generation. For an equitable evaluation, GAPS models trained exclusively on UniGeo CAL, UniGeo PRV, and PGPS9K datasets are used to assess the performance on their respective test datasets.}
\label{tab:hierarchical-beam-search}
\begin{tabular}{@{}lcc@{}}
\toprule
           & \multicolumn{1}{l}{Greedy Decode (\%)} & \multicolumn{1}{l}{Hierarchical Beam Search (\%)} \\ \midrule
CAL        & 42.1                              & \textbf{65.4} (\textit{+23.3})                                        \\ \midrule
PRV        & 97.5                              & \textbf{97.8} (\textit{+0.30})                                      \\ \midrule
PGPS9K     & 32.6                              & \textbf{50.5} (\textit{+17.9})                                \\ \midrule
Geometry3K & 35.8                              & \textbf{57.5} (\textit{+21.7})                                  \\ \bottomrule
\end{tabular}
\end{table}

Specifically, when utilizing the hierarchical beam search for solution program generation, GAPS achieves impressive accuracy improvements of 23.3\%, 0.3\%, 17.9\%, and 21.7\% on the UniGeo CAL, PRV, PGPS9K, and Geometry3K test data, respectively. This compelling evidence highlights the efficacy of the hierarchical beam search in allowing GAPS to effectively explore various combinations of operators and operands, ultimately leading to the selection of high-quality solution programs. The hierarchical beam search proves to be a crucial enhancement, enabling GAPS to achieve superior performance in generating accurate and diverse solution programs for geometry math problems.

\subsection{Enhancing Geometry Element Selection with the Geometry Elements Enhancement Method}

The geometry elements enhancement method significantly enhances GAPS' capability to select geometry elements as operands by appending the geometry elements to the problem text. In addition, this enhancement proves particularly valuable in situations where the elements involve identical points (e.g., \(\angle \mathit{SUW}\) and \(\triangle \mathit{SUM}\)), enabling GAPS to differentiate between such elements effectively. To substantiate the effectiveness of this approach, we compare the performances on UniGeo PRV problems between GAPS with the enhancement method and GAPS without it. The results, as presented in Table~\ref{tab:EPT}, clearly demonstrate the substantial benefits that GAPS gains from the geometry elements enhancement method.

\begin{table}[tbhp!]
\centering
\caption{Performance comparison between GAPS with (w) geometry elements enhancement and without (w/o) it.}
\begin{tabular}{ccccc}
\hline
        & \multicolumn{2}{c}{PRV (\%)} & \multicolumn{2}{c}{CAL (\%)} \\ \cmidrule(r){2-3} \cmidrule(r){4-5}
        & Top1        & Top10      & Top1        & Top10      \\ \hline
 w/o & 88.4       & 89.0     & 42.1      & 67.8     \\ \hline
 w   & 97.5       & 97.5      & 41.6     & 65.6     \\ \hline
\end{tabular}
\label{tab:EPT}
\end{table}

{We also undertake a case study where we map out the probability distributions during the generation of operands, as seen in Figure~\ref{fig:heat-map}. This instance is taken from the proof problem showcased in Figure~\ref{fig:geometry-problems}, where the intended solution program is designated as ``R\_4, E\_1, congruent, E\_3, R\_15, E\_3, similar, E\_2". Observing Figure~\ref{fig:heat-map}, it is evident that GAPS is inclined to choose geometry elements from the appended geometry elements while employing geometry element enhancement. Conversely, in the absence of geometry element enhancement, GAPS erroneously picks ``E\_0" (``triangle SUW") as the operand for the second sub-program, whereas the accurate operand should be ``E\_3" (``angle SUW"). We hypothesize this occurs due to the geometry elements being tokenized into multiple segments by the tokenizer, allowing these segments to easily engage with other contextual token pieces, subsequently generating extra noise. Nonetheless, appending these geometry elements to the problem statement can alleviate such noise, facilitating an enhancement in the precision of operand selection.

\begin{figure}[tbhp!] 
\centering
\includegraphics[width=0.8\textwidth]{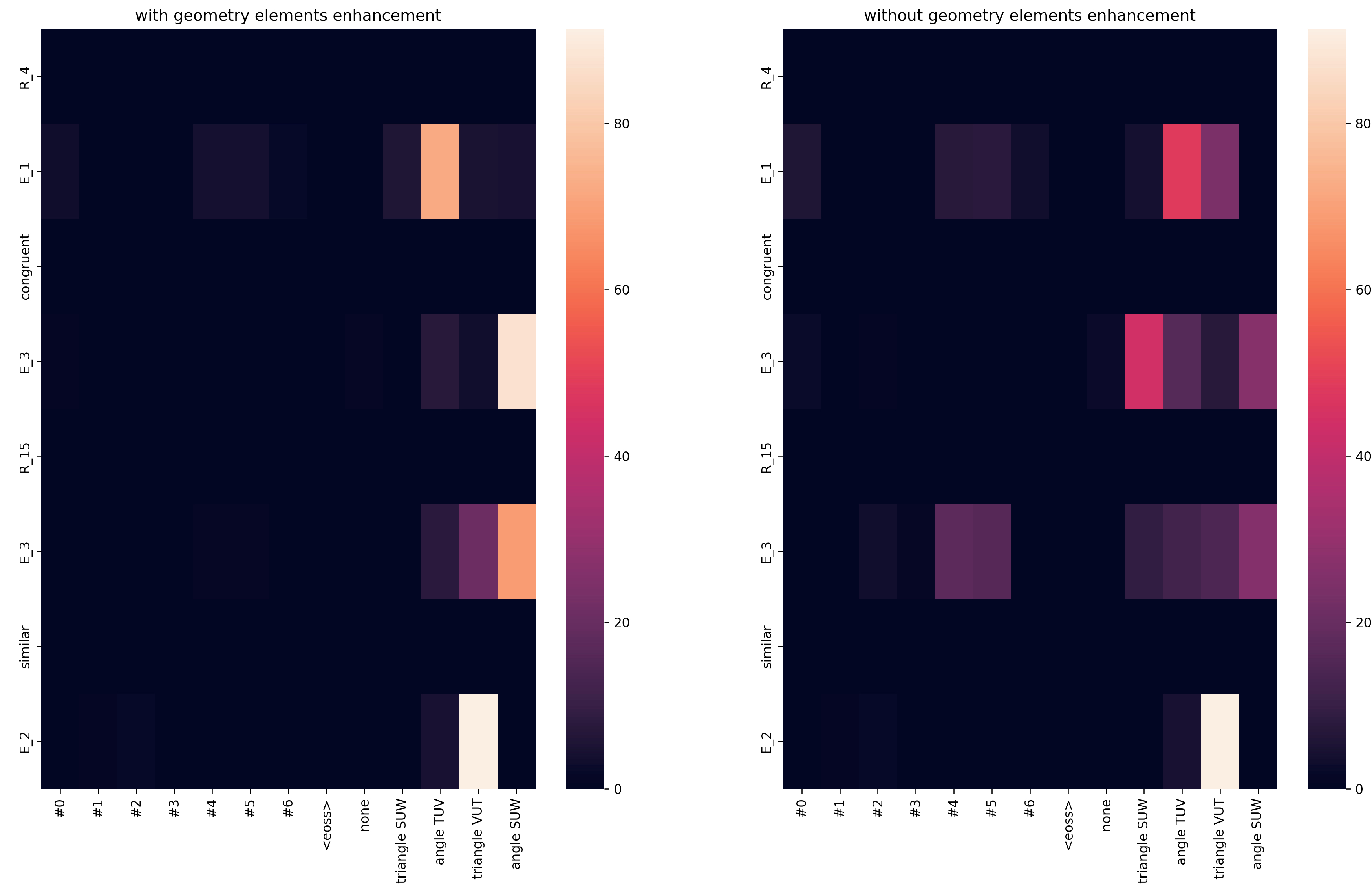} 
\caption{This heatmap illustrates the probability distributions across operand generation with and without the utilization of geometry element enhancement technology. It exemplifies the creation of a solution program for a geometry proving problem. We truncate the x-axis by omitting the constants not pertinent to solving proving problems. The y-axis denotes the tokens in the solution program, and the probabilities of generating operators are nullified, as this instance aims to scrutinize the impact of geometry element enhancement on the generation of operands. With the enhancement enabled, the value vectors for the geometry elements are derived from the vectors of the corresponding appended geometry elements in the problem text. Conversely, without the enhancement, the value vectors for the geometry elements are extracted from the vectors of the original geometry elements within the problem text.}
\label{fig:heat-map}
\end{figure}

Similarly, we explore the applicability of a method akin to the geometry elements enhancement to improve GAPS' performance on CAL problems. Since numbers in the text are not tokenized into unknown tokens, we solely insert the token "number" before each number and append all numbers to the problem text. However, as demonstrated in Table~\ref{tab:EPT}, contrary to the results observed with the geometry elements enhancement, GAPS' performances decline when employing this method. The findings in Table~\ref{tab:EPT} indicate that the approach of inserting "number" tokens and appending the numbers to the problem text does not yield the desired performance improvements for GAPS on CAL problems. Consequently, this method does not prove as effective in enhancing GAPS' ability to handle CAL problem types as compared to its success with PRV problems.

\section{Limitations and Future Work}

Indeed, while GAPS has achieved state-of-the-art results, there are still some limitations that we have identified in our model. One prominent limitation is that while GAPS performs impressively on geometry PRV problems, it does not exhibit the same level of performance on geometry CAL problems. We have conducted a thorough analysis and discovered that GAPS performs significantly better when the solution programs follow a specific pattern, where one operand follows each operator. This suggests a promising future direction for enhancing GAPS' performance on geometry CAL problems, by converting solution programs into a suitable pattern before employing GAPS.

Additionally, we introduced the geometry elements enhancement method, which significantly improves the model's accuracy in selecting geometry elements as operands for solution programs. However, during experimentation, we observed that applying a similar enhancement method for numbers negatively impacted the model's performance. This issue seems to stem from the challenge of designing meaningful indication tokens for numbers. In our future work, we aim to explore how performance improvements can be achieved specifically for geometry CAL problems, taking into account the nuances of handling numbers and designing appropriate indication tokens.

\section{Conclusion}

The automatic resolution of geometry math problems poses a desirable yet challenging task within the NLP community. To stimulate advancements in this domain, we propose the Geometry-Aware Problem Solver (GAPS) capable of simultaneously handling diverse types of geometry math problems. GAPS achieves this by integrating a problem-type classifier, enabling a unified solution program generator to cater to various problem types. This approach empowers GAPS to independently generate operators and operands by selecting symbols from domain-specific languages, specific to each problem type. Moreover, the problem-type classifier equips GAPS with the capacity to incorporate diverse types of data for supplementary training, without compromising performance. Our experimental results underscore this adaptability, illustrating that introducing new training data featuring distinct DSLs can notably enhance GAPS' proficiency in solving geometry math problems. Additionally, we introduce a hierarchical beam search technology that harmonizes with GAPS' paradigm of segregating operator and operand generation. This technology enables GAPS to select high-quality solution programs proficiently. Concluding our approach, we introduce a technique referred to as "geometry elements enhancement," aimed at bolstering GAPS' capability to discern geometry elements from the problem description provided. This enhancement method serves as a valuable tool for accurately identifying geometry elements within the context of geometry mathematical problems.

We conducted comprehensive experiments to assess the effectiveness of our GAPS model. Firstly, we compared GAPS against state-of-the-art (SOTA) models using the UniGeo dataset, and GAPS emerged as the clear winner, achieving state-of-the-art results. Notably, GAPS showcased its remarkable prowess by attaining an impressive 97.5\% accuracy on geometry PRV problems, highlighting its exceptional ability to tackle such challenges. The results underscore the superior capabilities of GAPS in solving geometry math problems, particularly when leveraging sophisticated fusion methods for precise and comprehensive reasoning in geometry math tasks. Furthermore, we have shown that training GAPS on a merged dataset featuring a variety of domain-specific languages (DSLs) can significantly boost its performance. This enhancement is made possible through the utilization of the unified solution program pattern and the problem-type classifier. Consequently, GAPS can proficiently tackle diverse types of geometry math problems with the utilization of a single solution program generator. Moreover, we thoroughly investigated the significance of the problem-type classifier and concluded that its benefits are maximized when accurately identifying the correct number of problem types. Additionally, we compared the performance of hierarchical beam search and greedy search, revealing the superiority of the hierarchical beam search in enabling GAPS to efficiently explore diverse combinations of operators and operands. Finally, we showcased the substantial benefits gained from the geometry elements enhancement method, which significantly improved GAPS' ability to recognize and handle geometry elements within the problem context. Overall, our findings provide strong evidence of GAPS' efficacy and the importance of incorporating advanced techniques to enhance its performance on geometry problem-solving tasks.



\bibliographystyle{elsarticle-num} 
\urlstyle{same} 
\bibliography{bibliography}




\end{document}